\documentclass[10pt,twocolumn,letterpaper]{article}

\usepackage[pagenumbers]{cvpr} 

\usepackage{multirow}
\usepackage[dvipsnames]{xcolor}
\usepackage{colortbl}  
\usepackage{subcaption}  
\usepackage{float}

\definecolor{cvprblue}{rgb}{0.21,0.49,0.74}
\usepackage[pagebackref,breaklinks,colorlinks,citecolor=cvprblue]{hyperref}

\def\paperID{23} 
\def\confName{3DV\xspace}
\def\confYear{2025\xspace}

\title{Semantic-Free Procedural 3D Shapes Are Surprisingly Good Teachers}

\author{Xuweiyi Chen \quad Zezhou Cheng \\
University of Virginia \\
\url{https://point-mae-zero.cs.virginia.edu/}
}
\begin{document}
\maketitle
\begin{abstract}
Self-supervised learning has emerged as a promising approach for acquiring transferable 3D representations from unlabeled 3D point clouds. Unlike 2D images, which are widely accessible, acquiring 3D assets requires specialized expertise or professional 3D scanning equipment, making it difficult to scale and raising copyright concerns. To address these challenges, we propose learning 3D representations from procedural 3D programs that automatically generate 3D shapes using simple 3D primitives and augmentations.

Remarkably, despite lacking semantic content, the 3D representations learned from the procedurally generated 3D shapes perform on par with state-of-the-art representations learned from semantically recognizable 3D models (\eg, airplanes) across various downstream 3D tasks, such as shape classification, part segmentation, masked point cloud completion, and both scene semantic and instance segmentation.
We provide a detailed analysis on factors that make a good 3D procedural programs.
Extensive experiments further suggest that current 3D self-supervised learning methods on point clouds do not rely on semantics of 3D shapes, shedding light on the nature of 3D representations learned. 
\vspace{-20pt}
\end{abstract}    
\section{Introduction}
\label{sec:intro}

Self-supervised learning (SSL) aims at learning representations from unlabeled data that can transfer effectively to various downstream tasks.
Inspired by the success of SSL in language~\cite{devlin2018bert} and 2D images~\cite{he2020momentum,he2022masked}, SSL for 3D point cloud understanding has gained considerable interest~\cite{wang2021occo,yu2022point,pang2022masked}. 
Recently, Point-MAE~\cite{pang2022masked} and its follow-ups~\cite{zhang2022point,zhang2024pcp,wu2023maskedscenecontrastscalable,wu2025sonata} exploit the masked autoencoding scheme~\cite{he2022masked} for 3D point cloud representation learning, showing substantial improvements in various 3D shape understanding tasks (\eg, shape classification, part segmentation, and scene instance segmentation).

However, unlike language and image data, which are abundantly available on the Internet, 3D assets are less accessible, as their creation often requires domain expertise and specialized tools such as 3D modeling software (\eg, Blender) or scanning equipment (\eg LiDAR sensors). 
This scarcity of 3D shapes, often referred to as the 3D data desert~\cite{dong2023autoencoderscrossmodalteacherspretrained}, has significantly hindered the scalability of representation learning methods. 
Recent efforts have expanded point cloud datasets at both the object level~\cite{deitke2023objaverse,deitke2024objaverse} and the scene level~\cite{baruch2021arkitscenes,fu20213d,yeshwanth2023scannet++}, but often rely on substantial human effort. 
Nevertheless, challenges unique to 3D data collection—such as copyright concerns, diverse file formats, and limited scalability—remain largely unresolved.

\begin{figure}[t!]
    \centering
    \includegraphics[width=\linewidth]{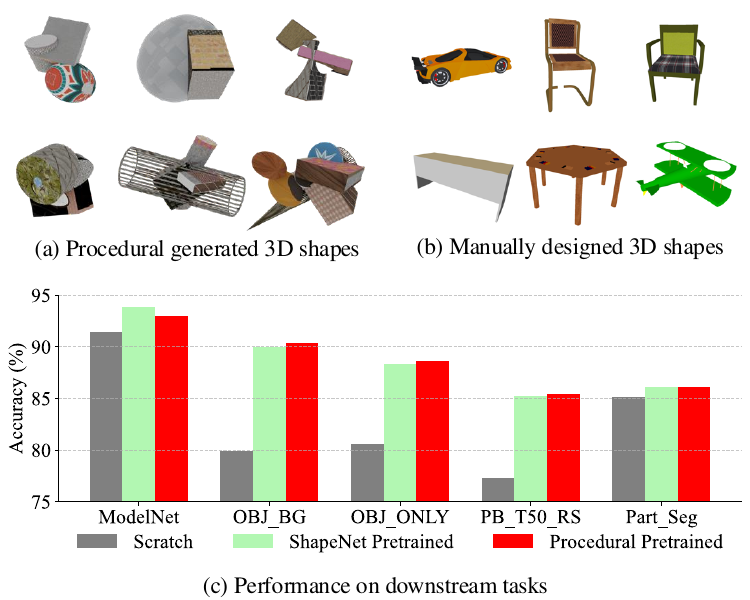}
     \vspace{-6mm}
    \caption{Self-supervised learning from \textbf{(a)} procedurally generated 3D shapes~\cite{xie2024lrm, xu2018deep, xu2019deep,jiang2024megasynth} performs comparably to learning from \textbf{(b)} ShapeNet models that are semantically meaningful~\cite{chang2015shapenet} across various downstream 3D understanding tasks. 
    Both outperforms training from scratch significantly.
    In \textbf{(c)}, the x-axis represents various tasks and benchmarks: ModelNet40~\cite{liu2019relation} and three variants of ScanObjectNN~\cite{uy2019revisiting} for shape classification, and ShapeNetPart~\cite{yi2016scalable} for part segmentation. \vspace{-10px}}
\label{fig:teaser}
\end{figure}

A common belief in 3D representation learning is that strong representations require semantically meaningful 3D shapes---objects such as chairs, airplanes, or indoor scenes---which are inherently costly to curate and difficult to scale. In this work, we challenge this assumption by asking a central question: \textbf{\emph{Do we really need semantically meaningful 3D shapes to learn strong 3D representations?}} 

To explore this, we investigate learning point cloud representations from \emph{purely synthetic data} generated by procedural 3D programs~\cite{xie2024lrm,xu2018deep,xu2019deep,jiang2024megasynth}, as illustrated in Fig.~\ref{fig:teaser}a. 
Our data generation pipeline begins by sampling simple 3D primitives (\eg, cubes, cylinders, spheres), which are then transformed via affine operations (\eg, scaling, translation, rotation) and composed into more complex geometries. 
We further apply augmentations such as Boolean operations to enrich topological diversity, followed by uniform surface point sampling to obtain point clouds suitable for representation learning. This approach is lightweight, efficient, and capable of generating unlimited number of 3D shapes with diverse geometric structures.
Unlike standard datasets, our procedural shapes lack human-recognizable semantics.

We then pose a follow-up question: \textbf{\emph{Are current 3D self-supervised learning (SSL) methods capable of leveraging such non-semantic, procedural shapes?}} 
To answer this, we generate 150K object-level and 4K scene-level synthetic 3D point clouds using our procedural programs, at a total cost of approximately 1400 CPU hours. 
Notably, the scale of our dataset exceeds that of widely used benchmarks such as ShapeNet (51K publicly available shapes) and ScanNet (1,513 indoor scenes).
We benchmark multiple representative 3D SSL methods, 
including Point-MAE~\cite{pang2022masked} and its recent variants~\cite{zhang2022point,zhang2024pcp,wu2023maskedscenecontrastscalable},
and evaluate them across a wide variety of downstream tasks such as shape classification, part segmentation, masked point cloud completion, and scene-level semantic and instance segmentation.

\noindent
Our main findings are as follows:
\begin{itemize}[itemsep=1pt, topsep=0.0pt]

\item \textbf{Semantic-free procedural 3D shapes are surprisingly good teachers.}
Despite lacking semantic content, self-supervised models trained solely on synthetic data perform on par with counterparts trained on real 3D datasets such as ShapeNet~\cite{chang2015shapenet}, Objaverse~\cite{deitke2023objaverse}, and ScanNet~\cite{dai2017scannet}. Moreover, they significantly outperform models trained from scratch without any pretraining (Fig.~\ref{fig:teaser}c, Tabs.~\ref{tab:fine-tuning}--\ref{tab:msc_results}).

\item \textbf{Geometric diversity plays a key role in learning effective 3D representations.}
We provide detailed insights into the factors that influence the quality of procedurally generated 3D datasets. Our analysis shows that learning performance improves significantly with increased geometric diversity and larger dataset size (Tab.~\ref{fig:dataset_complexity}, Fig.~\ref{fig:datasetsize}).

\item \textbf{Current 3D SSL methods rely more on geometric cues rather than semantic content.} 
Our in-depth analysis reveals strong structural similarities between representations learned from semantic-free synthetic procedural shapes and those learned from semantically meaningful 3D models (see t-SNE visualization in Fig.~\ref{fig:tsne}).
\end{itemize}

To our best knowledge, this is the first systematic large-scale study on 3D SSLs from procedural 3D shapes. 
Our work is inspired by recent works that successfully train large 3D reconstruction models exclusively on procedurally generated shapes~\cite{xie2024lrm, jiang2024megasynth}. 
Our exploration is also closely related to prior efforts that learn image or video representations from procedural programs~\cite{baradad2021learning, baradad2022procedural, yu2024learning}. 
While recent efforts focus on scaling 3D datasets using human-designed models or 3D scans~\cite{deitke2023objaverse,deitke2024objaverse,yeshwanth2023scannet++}, our approach is orthogonal and complementary, leveraging procedurally generated data to bypass manual design and scanning altogether.

\section{Related Work}
\label{sec:related}

\begin{figure*}[ht!]
    \centering
    \includegraphics[width=\linewidth]{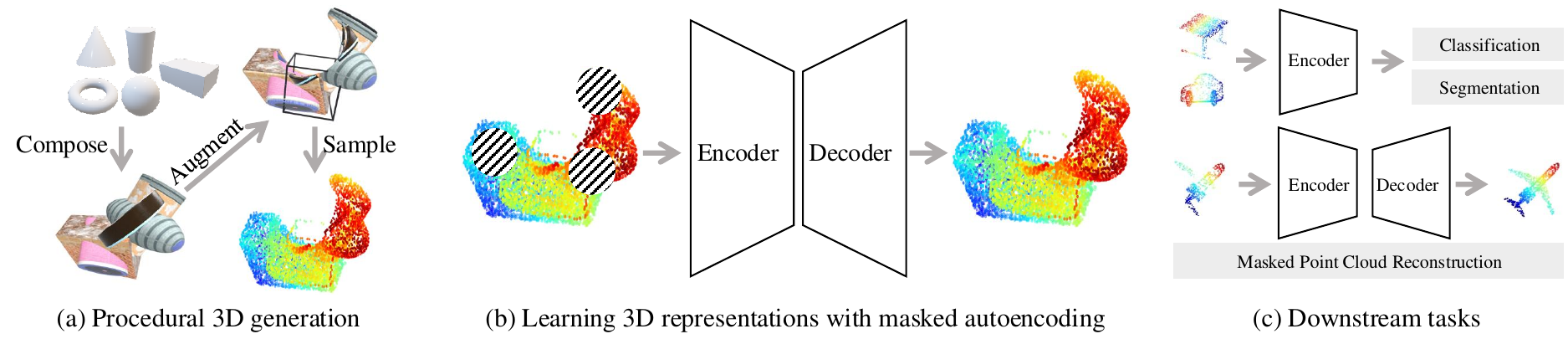}
    \vspace{-2mm}
    \caption{
    \textbf{Learning from procedural 3D programs.}
    \textbf{(a)} Synthetic 3D point clouds are generated by sampling, compositing, and augmenting simple primitives using procedural 3D programs~\cite{xie2024lrm}. \textbf{(b)} We experiment with multiple state-of-the-art self-supervised learning frameworks for learning 3D representations from synthetic data. Here, we illustrate the pretraining pipeline using Point-MAE~\cite{yu2022point}, naming this variant \emph{Point-MAE-Zero}, where ``Zero" emphasizes the absence of any human-made 3D shapes. \textbf{(c)} We evaluate the pretrained models across various 3D shape understanding tasks.
    \vspace{-10px}
    }
\label{fig:pipeline}
\end{figure*}

\noindent
\textbf{3D Datasets.}
Significant efforts have been made to curate extensive 3D shape datasets~\cite{chang2015shapenet,modelnet40,reizenstein2021common,sun2018pix3d,collins2022abo,fu20213d,ma2024imagenet3d,wu2023omniobject3d,deitke2023objaverse}. For example, ShapeNet provides 3 million CAD models, with 51K high-quality shapes publicly available.
More recently, Objaverse~\cite{deitke2023objaverse} and Objaverse-XL~\cite{deitke2024objaverse} expanded the 3D dataset to 10.2 million manually-created 3D shapes. 
However, challenges, such as format diversity and copyright and legal issues, remain unsolved. 
At the scene level, most commonly used datasets—ScanNet~\cite{dai2017scannet}, Structured3D~\cite{structured3d}, Matterport3D~\cite{matterport3d}, nuScenes~\cite{nuscenes2019}, SemanticKITTI~\cite{semantickitti}, and Waymo~\cite{waymo}—are distributed under non-commercial or research-only licenses, limiting their applicability for broader use. A more recent effort, ASE~\cite{ASE}, introduces 100,000 synthetic indoor scenes, also under a non-commercial license.
In contrast, we explore procedural 3D programs that generate shapes and scenes from simple primitives, enabling the creation of unlimited synthetic objects and scenes without licensing constraints.

\noindent
\textbf{Learning from Synthetic Data.}
Synthetic data has become popular in computer vision, especially in scenarios where ground-truth annotations are difficult to obtain or where privacy and copyright issues arise. State-of-the-art performance in mid-level or 3D vision tasks is often achieved through training on synthetic data, including tasks like optical flow~\cite{teed2020raft}, depth estimation~\cite{yang2024depth}, dense tracking~\cite{karaev2023cotracker}, relighting~\cite{xu2018deep}, novel view synthesis~\cite{xu2019deep}, and material estimation~\cite{li2018learning}.
Procedurally generated synthetic data has also been explored for self-supervised representation learning in images~\cite{baradad2021learning,baradad2022procedural} and videos~\cite{yu2024learning}, and more recently for multi-view feed-forward 3D reconstruction~\cite{xie2024lrm}. 
In this work, we explore self-supervised representation learning for point clouds, using synthetic 3D shapes generated by procedural 3D programs.

\noindent
\textbf{Self-supervised Learning for Point Clouds.}
Recent self-supervised learning (SSL) methods for point clouds generally fall into two categories: contrastive learning~\cite{he2020momentum,xie2020pointcontrast,chhipa2022depth,gadelha2020label} and masked autoencoding~\cite{wang2021occo,yu2022point,pang2022masked,dong2023autoencoderscrossmodalteacherspretrained,qi2023recon,guo2023jointmae2d3djointmasked,zhang2022learning3drepresentations2d,zha2023compact3drepresentationspoint,zha2023towards}. 
Contrastive approaches such as PointContrast~\cite{xie2020pointcontrast} and DepthContrast~\cite{chhipa2022depth} rely on instance discrimination~\cite{he2020momentum} to learn view-invariant features. 
Inspired by the success of masked modeling in vision~\cite{he2022masked} and language~\cite{devlin2018bert}, Point-BERT~\cite{yu2022point} and Point-MAE~\cite{pang2022masked} use transformer-based architectures to predict masked regions of the point cloud. 
Point-M2AE~\cite{zhang2022point} extends Point-MAE with a multi-scale pyramid design, while PCP-MAE~\cite{zhang2024pcp} addresses centroid leakage by adding centroid prediction as an auxiliary objective. 
Scene-level SSL methods that operate directly on large-scale point clouds have also gained traction~\cite{wu2023maskedscenecontrastscalable,hou2021exploringdataefficient3dscene,xie2020pointcontrastunsupervisedpretraining3d,Wang_2024_CVPR,liu2024pointcloudunsupervisedpretraining}.
In this work, we adopt Point-MAE, Point-M2AE, PCP-MAE, and MSC as our primary SSL frameworks due to their strong performance on downstream 3D tasks. 
For ablation studies, we use Point-MAE as our baseline, as it represents a foundational and widely adopted approach in this line of research.
\section{Learning from Procedural 3D Programs}
\label{sec:method}

We first introduce the procedural 3D programs~\cite{xie2024lrm,xu2018deep,xu2019deep} for generating unlimited number of synthetic 3D shapes using composition of simple primitive shapes (\eg, cylinders) and  shape augmentation (Sec.~\ref{sec:program}). We then describe the masked autoencoding scheme~\cite{wang2021occo,yu2022point,pang2022masked} for learning 3D representations from synthetic 3D datasets (Sec.~\ref{sec:point-mae-zero}). 

\subsection{Procedural 3D programs}
\label{sec:program}
There is a line of work synthesizing procedural 3D shapes for vision tasks such as novel view synthesis~\cite{xu2019deep}, relighting~\cite{xu2018deep}, and material estimation~\cite{li2018learning}. Following recent methods~\cite{xie2024lrm} that use procedural 3D shapes for sparse-view reconstruction, we address self-supervised 3D representation learning from purely synthetic datasets. Fig.~\ref{fig:pipeline}a illustrates our data pipeline: 

\vspace{-10pt}
\begin{itemize}[label={}, itemsep=2pt, topsep=8pt, leftmargin=2pt]
    \item \textbf{(1)} Randomly sample $\mathbf{K}$ primitive shapes (cubes, spheres, cylinders, cones, tori) and apply affine transformations to combine them;
    \item \textbf{(2)} Apply geometric augmentations (e.g., boolean differences, wireframe conversions) to enrich shape diversity (see~\cite{xie2024lrm} for details);
    \item \textbf{(3)} Uniformly sample $\mathbf{N}$ surface points per synthesized shape as inputs for representation learning (Fig.~\ref{fig:pipeline}b).
\end{itemize}
\vspace{-10pt}

We experiment with various shape-generation configurations, such as changing the number of sampled primitives and applying augmentations. By default, each dataset consists of 150K shapes with $\mathbf{N}=8192$ points each. Sec.~\ref{sec:exp} further analyzes the effects of dataset size and shape complexity on learned representations.

In order to generate procedural 3D scenes, we follow MegaSynth~\cite{jiang2024megasynth} to procedurally generate 4K synthetic 3D scenes. Specifically, we first generate a floor plan and generate procedural 3D shapes with the above pipeline and place procedural 3D shapes in the scene based on the generated floor plan. We provide details on the generation of 4K procedural 3D scenes and visualizations of the generated shapes in the supplementary material.

\subsection{Procedural Pretraining}
\label{sec:point-mae-zero}

\paragraph{Pretraining.} 
We adopt Point-MAE~\cite{he2022masked}, Point-M2AE~\cite{zhang2022point}, PCP-MAE~\cite{zhang2024pcp} to train on procedural 3D shapes (Fig.~\ref{fig:pipeline}b). These methods rely on a masked autoencoding scheme~\cite{devlin2018bert,he2022masked,yu2022point}, where the input point cloud is split into irregular patches and a large portion of them (60\% by default) is randomly masked. A Transformer-based encoder--decoder network then attempts to reconstruct these masked patches, thereby learning 3D representations.
The reconstruction loss is computed as the $L_2$ Chamfer Distance between the predicted point patches $P_{\text{pre}}$ and the ground-truth patches $P_{\text{gt}}$:
\begin{align}
\label{eq:recon}
L = \sum_{x \in \{P_{\text{pre}}, P_{\text{gt}}\}} \frac{1}{|x|} \sum_{a \in x} \min_{b \in x'} \|a - b\|_2^2
\end{align}
where  $x'=P_{\text{gt}}$ if $x=P_{\text{pre}}$, and vice versa.
For scene-level SSLs, we adopt MSC~\cite{wu2023maskedscenecontrastscalable}, which combines masked auto-encoding and contrasive learning, to train on procedurally generated 3D scenes.

\vspace{-10px}
\paragraph{Downstream Probing.}
We evaluate baselines on several 3D tasks, as summarized in Fig.~\ref{fig:pipeline}c. For shape classification, we augment the pretrained Transformer encoder with a three-layer MLP classification head. For part segmentation, we aggregate features from the 4th, 8th, and final layers of the encoder, upsample them to all 2048 input points, and employ a segmentation head. For masked point cloud reconstruction, we use both the pretrained encoder and decoder with no architectural modifications. For scene-level methods, we use both instance segmentation and semantic segmentation finetuned from the pretrained SSLs. Detailed implementation settings are in the supplementary material.
\section{Experiments}
\label{sec:exp}

We present a comprehensive evaluation of 3D shape representations pretrained with procedural 3D programs across various downstream object-level and scene-level tasks, including object classification, part segmentation, and 3D scene understanding (Sec.~\ref{sec:cls} --~\ref{sec:scene}). 
We further provide an in-depth analysis of model behavior and ablation studies (Sec.~\ref{sec:analysis}). 
For each downstream task, we report the performance of relevant existing methods as a reference and focus on comparisons with SSLs pretrained on manually-curated 3D datasets, as well as models trained from scratch. 
Specifically, for object-level 3D understanding tasks, we evaluate the following three pretraining strategies: (1) \emph{Scratch}: All network parameters are randomly initialized, with no pretraining. (2) \emph{ShapeNet Pretrained (SN)}: Pretrained on 41,952 models in the ShapeNet~\cite{chang2015shapenet} training split, relying on the officially released weights. 
(3) \emph{Procedural 3D Programs Pretrained (Zero)}: Pretrained on 150K procedurally generated 3D models, using no human-crafted shapes.
For scene-level tasks, we compare SSL models pretrained on ScanNet~\cite{dai2017scannet} and procedurally generated 3D scenes.

\subsection{Object Classification}
\label{sec:cls}

\begin{table}
\resizebox{\columnwidth}{!}{
  \centering
  \begin{tabular}{@{} l c c c c | l@{}}
    \toprule
    Methods  & ModelNet40 & OBJ-BG & OBJ-ONLY & PB-T50-RS & Avg. \\
    \midrule
    PointNet~\cite{qi2017pointnet}  & 89.2 & 73.3 & 79.2 & 68.0 & 77.4 \\
    SpiderCNN~\cite{xu2018spidercnn} & 92.4 & 77.1 & 79.5 & 73.7 & 80.7 \\
    PointNet++~\cite{qi2017pointnet++} & 90.7 & 82.3 & 84.3 & 77.9 & 83.8 \\
    DGCNN~\cite{wang2019dynamic} & 92.9 & 86.1 & 85.5 & 78.5 & 85.8 \\
    PointCNN~\cite{li2018pointcnn}  & -- & 86.1 & 85.5 & 78.5 & -- \\
    PTv1~\cite{zhao2021point} & 93.7 & -- & -- & -- & -- \\
    PTv2~\cite{wu2022point} & 94.2 & -- & -- & -- & -- \\
    OcCo~\cite{wang2021occo} & 92.1 & 84.9 & 85.5 & 78.8 & 85.3 \\
    Point-BERT~\cite{yu2022point} & 93.2 & 87.4 & 88.1 & 83.1 & 88.0 \\
    \midrule
    Point-MAE-Scratch~\cite{yu2022point} & 91.4 & 79.9 & 80.6 & 77.2 & 82.3 \\
    \rowcolor{gray!10} Point-MAE-SN~\cite{pang2022masked} & \textbf{93.8} & 90.0 & 88.3 & 85.2 & 89.3 \textsubscript{\textcolor{red}{(+7.0)}} \\
    \rowcolor{blue!20} Point-MAE-Zero & 93.0 & \textbf{90.4} & \textbf{88.6} & \textbf{85.5} & 89.4 \textsubscript{\textcolor{red}{(+7.1)}} \\
    \midrule
    Point-M2AE-Scratch & 92.2 & 90.0 & 87.6 & 85.6 & 88.9 \\
    \rowcolor{gray!10} Point-M2AE-SN~\cite{zhang2022point} & \textbf{94.0} & \textbf{91.2} & 88.8 & 86.4 & 90.1 \textsubscript{\textcolor{red}{(+1.2)}} \\
    \rowcolor{blue!20} Point-M2AE-Zero & 92.9 & 90.4 & \textbf{89.8} & \textbf{87.0} & 90.0 \textsubscript{\textcolor{red}{(+1.1)}} \\
    \midrule
    PCP-MAE-Scratch & 91.5 & 88.8 & 88.5 & 83.8 & 88.2 \\
    \rowcolor{gray!10} PCP-MAE-SN~\cite{zhang2024pcp} & \textbf{94.0} & \textbf{95.5} & \textbf{94.3} & 90.4 & 93.6 \textsubscript{\textcolor{red}{(+5.4)}} \\
    \rowcolor{blue!20} PCP-MAE-Zero & 92.4 & 94.0 & 92.3 & \textbf{90.5} & 92.3 \textsubscript{\textcolor{red}{(+4.1)}} \\
    \bottomrule
  \end{tabular}
}
\vspace{-2mm}
\caption{\textbf{Object Classification.} We evaluate the object classification performance on ModelNet40 and three variants of ScanObjectNN. Classification accuracy (\%) is reported (\emph{higher is better}). \textbf{Top}: Performance of existing methods with various neural network architectures and pretraining strategies. \textbf{Bottom}: Comparison with our baseline methods. The rightmost column shows the average accuracy, with red text indicating the improvement over the corresponding Scratch baseline.\vspace{-10px}}
\label{tab:fine-tuning}
\end{table}

\paragraph{Benchmarks.} 
We use ModelNet40~\cite{wu20153d} and ScanObjectNN~\cite{uy2019revisiting} as the benchmarks for the shape classification task. 
ModelNet40 contains 12,311 clean 3D CAD objects across 40 categories, with 9,843 samples for training and 2,468 for testing. Following Point-MAE, we apply random scaling and translation as data augmentation during training, and a voting strategy during testing~\cite{liu2019relation}.
Following prior works~\cite{wang2021occo,yu2022point,pang2022masked}, we also evaluate the few-shot classification performance on ModelNet40. 
ScanObjectNN is a more complex real-world 3D dataset, consisting of approximately 15,000 objects across 15 categories, with items scanned from cluttered indoor scenes. We report results on three ScanObjectNN variants: OBJ-BG, OBJ-ONLY, and PB-T50-RS, the latter being the \textit{most challenging} due to its additional noise and occlusions.

\vspace{-5pt}
\paragraph{Transfer Learning.}
Table~\ref{tab:fine-tuning} summarizes object classification results across several settings. 
On ModelNet40, the ``-Zero'' variants (\eg, Point-MAE-Zero, Point-M2AE-Zero, PCP-MAE-Zero) generally fall slightly behind their ShapeNet‐pretrained counterparts (``-SN''), reflecting the larger domain gap between synthetic shapes and the clean 3D models in ModelNet40. 
By contrast, on ScanObjectNN---which contains real-world scans with broader geometric variability---the ``-Zero'' models often match or exceed the performance of their ``-SN'' counterparts. 
For instance, PCP-MAE-Zero outperforms PCP-MAE-SN on the PB-T50-RS variant, and Point-M2AE-Zero closely matches or exceeds Point-M2AE-SN in several cases. 
These findings indicate that the diverse geometry in procedurally synthesized data can be advantageous for certain real-world tasks. 
Meanwhile, all pretrained models (including both ``-SN'' and ``-Zero'') surpass their respective from-scratch baselines and outperform existing self-supervised approaches~\cite{wang2021occo,yu2022point} which we highlight with the rightmost column.

\begin{table}[t]
\resizebox{\columnwidth}{!}{
  \centering
  \begin{tabular}{@{}l c c c c | l@{}}
    \toprule
    Methods & 5w/10s & 5w/20s & 10w/10s & 10w/20s & Avg. \\
    \midrule
    DGCNN-rand~\cite{wang2019dynamic} & 31.6$\pm$2.8 & 40.8$\pm$4.6 & 19.9$\pm$2.1 & 16.9$\pm$1.5 & 27.3 \\
    DGCNN-OcCo~\cite{wang2019dynamic} & 90.6$\pm$2.8 & 92.5$\pm$1.9 & 82.9$\pm$1.3 & 86.5$\pm$2.2 & 88.1 \\
    Transformer-OcCo~\cite{wang2021occo} & 94.0$\pm$3.6 & 95.9$\pm$2.3 & 89.4$\pm$5.1 & 92.4$\pm$4.6 & 92.9 \\
    Point-BERT~\cite{yu2022point} & 94.6$\pm$3.1 & 96.3$\pm$2.7 & 91.0$\pm$5.4 & 92.7$\pm$5.1 & 93.7 \\
    \midrule
    Point-MAE-Scratch~\cite{yu2022point} & 87.8$\pm$5.2 & 93.3$\pm$4.3 & 84.6$\pm$5.5 & 89.4$\pm$6.3 & 88.8 \\
    \rowcolor{gray!10} Point-MAE-SN~\cite{pang2022masked} & 96.3$\pm$2.5 & \textbf{97.8$\pm$1.8} & \textbf{92.6$\pm$4.1} & 95.0$\pm$3.0 & 95.4~\textsubscript{\textcolor{red}{(+6.6)}} \\
    \rowcolor{blue!20} Point-MAE-Zero & \textbf{96.6$\pm$2.2} & 97.6$\pm$1.4 & 91.9$\pm$4.2 & \textbf{95.2$\pm$3.0} & 95.3~\textsubscript{\textcolor{red}{(+6.5)}} \\
    \midrule 
    Point-M2AE-Scratch & 87.5$\pm$2.6 & 90.0$\pm$5.5 & 86.4$\pm$3.2 & 89.6$\pm$4.3 & 88.4 \\
    \rowcolor{gray!10} Point-M2AE-SN$^{*}$~\cite{pang2022masked} & 93.4$\pm$3.1 & \textbf{96.2$\pm$1.5} & 91.8$\pm$4.5 & 92.9$\pm$3.2 & 93.6~\textsubscript{\textcolor{red}{(+5.2)}} \\
    \rowcolor{blue!20} Point-M2AE-Zero & \textbf{95.4$\pm$2.8} & 94.4$\pm$2.8 & \textbf{94.3$\pm$2.2} & \textbf{93.8$\pm$3.2} & 94.5~\textsubscript{\textcolor{red}{(+6.1)}} \\
    \midrule
    PCP-MAE-Scratch & 86.4$\pm$2.6 & 85.0$\pm$6.0 & 88.9$\pm$4.1 & 90.7$\pm$4.2 & 87.8 \\
    \rowcolor{gray!10} PCP-MAE-SN~\cite{zhang2024pcp} & \textbf{97.4$\pm$2.3} & \textbf{99.1$\pm$0.8} & 93.5$\pm$3.7 & \textbf{95.9$\pm$2.7} & 96.5~\textsubscript{\textcolor{red}{(+8.7)}} \\
    \rowcolor{blue!20} PCP-MAE-Zero & 95.5$\pm$3.4 & 98.6$\pm$1.6 & \textbf{94.2$\pm$3.5} & 95.6$\pm$3.0 & 96.0~\textsubscript{\textcolor{red}{(+8.2)}} \\
    \bottomrule
  \end{tabular}
}
\caption{\textbf{Few-shot classification on ModelNet40.} We evaluate performance on four \(n\)-way, \(m\)-shot configurations. For example, 5w/10s denotes a 5-way, 10-shot classification task. The table reports the mean classification accuracy (\%) and standard deviation across 10 independent runs for each configuration. \textbf{Top}: Results from existing methods for comparison. \textbf{Bottom}: Comparison with our baseline methods. Note that results for Point-M2AE-SN are reproduced using publicly available code, as the original configuration was not provided. The final column shows the average accuracy across configurations, with subscripts indicating improvements over the corresponding baseline. \vspace{-15px}}
\label{tab:fewshot}
\end{table}

\vspace{-15px}
\paragraph{Few-shot Classification.}
We evaluate few-shot classification on ModelNet40 using standard \(n\)-way, \(m\)-shot protocols, where \(n\) denotes the number of randomly selected classes and \(m\) the number of examples per class. Each evaluation samples 20 unseen instances from each class. We repeat this procedure 10 times, reporting mean accuracy (\%) and standard deviation.
Table~\ref{tab:fewshot} presents results for \(n=\{5,10\}\) and \(m=\{10,20\}\). Similar to transfer learning experiments, Point-MAE-Zero performs on par or slightly below Point-MAE-SN, likely due to the larger domain gap between procedural shapes and ModelNet40 data. Nonetheless, both methods substantially outperform their scratch-trained counterparts, as reflected by the performance deltas, and also surpass prior approaches such as DGCNN~\cite{wang2019dynamic} and Transformer-OcCo~\cite{wang2021occo}. 

\subsection{Part Segmentation} 
\label{sec:seg}

The 3D part segmentation task aims to assign a part label to each point in a shape. We evaluate our methods and baselines on ShapeNetPart~\cite{yi2016scalable}, which contains 16,881 models across 16 object categories. Consistent with previous works~\cite{qi2017pointnet,pang2022masked,yu2022point}, we sample 2,048 points from each shape, resulting in 128 patches in our masked autoencoding pipeline (see Sec.~\ref{sec:method}).

Table~\ref{tab:partseg} presents the mean Intersection-over-Union (mIoU) across all instances, along with per-category IoU. Across various models, both Point-MAE-Zero and Point-MAE-SN deliver comparable performance, indicating that procedurally generated shapes can learn robust 3D representations without explicit semantic content. Similarly, Point-M2AE-Zero and PCP-MAE-Zero achieve results on par with their ShapeNet-pretrained counterparts, further highlighting the versatility of procedural data in self-supervised representation learning. 

In line with our observations in Sec.~\ref{sec:cls}, the ``-Zero" and ``-SN" models surpass scratch-trained baselines and earlier methods that use different architectures~\cite{qi2017pointnet,qi2017pointnet++,wang2019dynamic} or alternative pretraining strategies~\cite{wang2021occo,yu2022point}. Despite lacking high-level semantic cues, these procedurally trained autoencoders still capture sufficient geometric structure to achieve strong segmentation performance.

\begin{table}[t]
\centering
\resizebox{\columnwidth}{!}{
  \begin{tabular}{@{}l | l | c c c c c c c c@{}}
    \toprule
    Methods & mIoU\textsubscript{I} & aero & bag & cap & car & chair & earphone & guitar & knife \\
    \midrule
    PointNet~\cite{qi2017pointnet} & 83.7 & 83.4 & 78.7 & 82.5 & 74.9 & 89.6 & 73.0 & 91.5 & 85.9 \\
    PointNet++~\cite{qi2017pointnet++} & 85.1 & 82.4 & 79.0 & 87.7 & 77.3 & 90.8 & 71.8 & 91.0 & 85.9 \\
    DGCNN~\cite{wang2019dynamic} & 85.2 & 84.0 & 83.4 & 86.7 & 77.8 & 90.6 & 74.7 & 91.2 & 87.5 \\
    OcCo~\cite{wang2021occo} & 85.1 & 83.3 & 85.2 & 88.3 & 79.9 & 90.7 & 74.1 & 91.9 & 87.6 \\
    Point-BERT~\cite{yu2022point} & 85.6 & 84.3 & 84.8 & 88.0 & 79.8 & 91.0 & 81.7 & 91.6 & 87.9 \\
    \midrule
    Point-MAE-Scratch~\cite{yu2022point} & 85.1 & 82.9 & \textbf{85.4} & 87.7 & 78.8 & 90.5 & \textbf{80.8} & 91.1 & \textbf{87.7} \\
    \rowcolor{gray!10} Point-MAE-SN~\cite{pang2022masked} & \textbf{86.1}\textsubscript{\textcolor{red}{(+1.0)}} & 84.3 & 85.0 & 88.3 & 80.5 & 91.3 & 78.5 & \textbf{92.1} & 87.4 \\  
    \rowcolor{blue!20} Point-MAE-Zero & \textbf{86.1}\textsubscript{\textcolor{red}{(+1.0)}} & \textbf{85.0} & 84.2 & \textbf{88.9} & \textbf{81.5} & \textbf{91.6} & 76.9 & \textbf{92.1} & 87.6 \\
    \midrule
    Point-M2AE-Scratch & 84.7 & 85.1 & 86.8 & 88.6 & \textbf{81.1} & 91.5 & 79.9 & \textbf{92.1} & 87.8 \\  
    \rowcolor{gray!10} Point-M2AE-SN~\cite{zhang2022point} & \textbf{85.0}\textsubscript{\textcolor{red}{(+0.3)}} & 84.5& 87.2 & \textbf{89.3} & \textbf{81.1} & \textbf{91.8} & \textbf{80.1} & 92.0 & \textbf{89.2} \\  
    \rowcolor{blue!20} Point-M2AE-Zero & 84.9\textsubscript{\textcolor{red}{(+0.2)}}& \textbf{85.3} & \textbf{87.3} & 88.7 & \textbf{81.1} & 91.7 & 79.4 & 91.9 & 88.2 \\
    \midrule
     PCP-MAE-Scratch & 83.8 & 84.3 & 83.1 & 88.7 & 80.3 & 91.2 & 77.1 & 92.0 & \textbf{88.1} \\  
    \rowcolor{gray!10} PCP-MAE-SN~\cite{zhang2024pcp} & 84.3\textsubscript{\textcolor{red}{(+0.5)}} & \textbf{85.0} & 84.0 & \textbf{88.7} & 81.0 & \textbf{91.6} & 77.6 & 91.8 & 87.6 \\  
    \rowcolor{blue!20} PCP-MAE-Zero & \textbf{84.4}\textsubscript{\textcolor{red}{(+0.6)}} & 84.6 & \textbf{84.3} & 88.5 & \textbf{81.7} & 91.5 & \textbf{81.1} & \textbf{92.1} & 87.0 \\
    \bottomrule
  \end{tabular}
}

\vspace{2mm}

\resizebox{\columnwidth}{!}{
  \begin{tabular}{@{}l | c c c c c c c c c@{}}
    \toprule
    Methods & lamp & laptop & motor & mug & pistol & rocket & skateboard & table \\
    \midrule
    PointNet~\cite{qi2017pointnet} & 80.8 & 95.3 & 65.2 & 93.0 & 81.2 & 57.9 & 72.8 & 80.6 \\
    PointNet++~\cite{qi2017pointnet++} & 83.7 & 95.3 & 71.6 & 94.1 & 81.3 & 58.7 & 76.4 & 82.6 \\
    DGCNN~\cite{wang2019dynamic} & 82.8 & 95.7 & 66.3 & 94.9 & 81.1 & 63.5 & 74.5 & 82.6 \\
    OcCo~\cite{wang2021occo} & 84.7 & 95.4 & 75.5 & 94.4 & 84.1 & 63.1 & 75.7 & 80.8 \\ 
    Point-BERT~\cite{yu2022point} & 85.2 & 95.6 & 75.6 & 94.7 & 84.3 & 63.4 & 76.3 & 81.5 \\ 
    \midrule
    Point-MAE-Scratch~\cite{yu2022point} & 85.3 & 95.6 & 73.9 & \textbf{94.9} & 83.5 & 61.2 & 74.9 & 80.6 \\
    \rowcolor{gray!10} Point-MAE-SN~\cite{pang2022masked} & \textbf{86.1} & \textbf{96.1} & 75.2 & 94.6 & 84.7 & 63.5 & 77.1 & \textbf{82.4} \\  
    \rowcolor{blue!20} Point-MAE-Zero & 86.0 & 96.0 & \textbf{77.8} & 94.8 & \textbf{85.3} & \textbf{64.7} & \textbf{77.3} & 81.4 \\
    \midrule
    Point-M2AE-Scratch & 85.7 & 96.0 & 76.4 & \textbf{95.4} & \textbf{85.5} & 63.8 & 76.3 & 82.4 \\  
    \rowcolor{gray!10} Point-M2AE-SN~\cite{zhang2022point} & \textbf{86.4} & 95.8 & \textbf{77.7} & 95.3 & 85.2 & \textbf{65.3} & \textbf{77.0} & 82.2 \\  
    \rowcolor{blue!20} Point-M2AE-Zero & 85.8 & \textbf{96.2} & 76.6 & 
    94.9 & 84.8 & 64.4 & 76.8 & \textbf{82.5} \\
    \midrule
    PCP-MAE-Scratch & 84.9 & 95.0 & 76.0 & 95.0 & 85.0 & 63.2 & 75.4 & 81.0 \\  
    \rowcolor{gray!10} PCP-MAE-SN~\cite{zhang2024pcp} & 85.8 & \textbf{96.4} & 76.1 & \textbf{95.2} & 84.8 & \textbf{64.0} & \textbf{77.4} & \textbf{81.4} \\  
    \rowcolor{blue!20} PCP-MAE-Zero & \textbf{86.0} & 96.1 & \textbf{76.6} & 94.6 & \textbf{85.1} & 63.6 & 76.8 & 80.4 \\
    \bottomrule
  \end{tabular}
}
\caption{\textbf{Part Segmentation Results.} We report the mean Intersection over Union over instances (mIoU\textsubscript{I}) and the per-category IoU (\%) on the ShapeNetPart benchmark. Higher values indicate better performance. The subscripted values in mIoU\textsubscript{I} represent performance improvement over the corresponding baseline.}
\label{tab:partseg}
\end{table}
\begin{figure*}[t!]
    \centering
        \includegraphics[width=\linewidth]{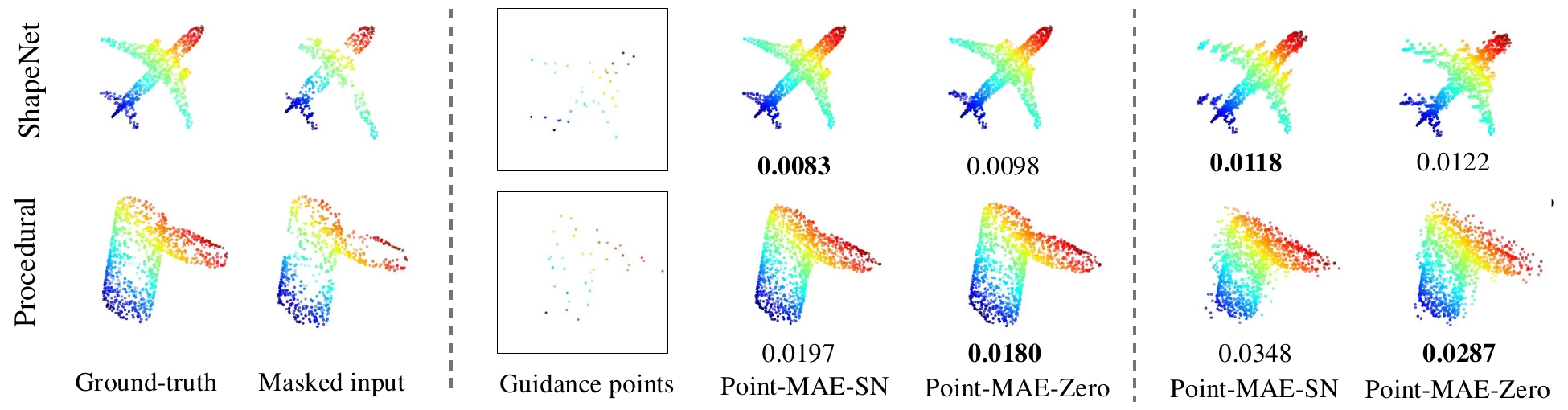}
    \vspace{-2mm}
    \caption{\textbf{Masked Point Cloud Completion.} This figure visualizes shape completion results with Point-MAE-SN and Point-MAE-Zero on the ShapeNet test split and procedurally synthesized 3D shapes.
    \textbf{Left}: Ground truth point clouds and masked inputs (60\% mask ratio).
    \textbf{Middle}: Completions guided by masked input patch centers~\cite{pang2022masked}.
    \textbf{Right}: Reconstructions without any guidance points. 
    The $L_2$ Chamfer distance (\emph{lower is better}) between the predicted 3D point clouds and the ground truth is displayed below each reconstruction.
    \vspace{-5pt}
}
    \label{fig:recon}
\end{figure*}

\begin{table}[t]
\resizebox{\columnwidth}{!}{
  \centering
  \begin{tabular}{@{}l c c c c@{}}
    \toprule
    & \multicolumn{2}{c}{With Guidance} & \multicolumn{2}{c}{Without Guidance} \\
     \cmidrule(lr){2-3} \cmidrule(lr){4-5}
    Methods  & ShapeNet & Synthetic & ShapeNet & Synthetic\\
    \midrule
    Point-MAE-SN~\cite{pang2022masked} & \textbf{0.015} & \textbf{0.024} & \textbf{0.024} & 0.039\\
    \rowcolor{blue!20} Point-MAE-Zero & 0.016 & \textbf{0.024} & 0.026 & \textbf{0.037} \\
    \midrule
    Point-M2AE-SN~\cite{zhang2022point} & \textbf{0.002} & \textbf{0.005} & \textbf{0.007} & 0.011 \\
    \rowcolor{blue!20} Point-M2AE-Zero & 0.003 & \textbf{0.005} &  0.010 & \textbf{0.009} \\
    \midrule
    PCP-MAE-SN~\cite{zhang2024pcp} & - & - & \textbf{0.016} & \textbf{0.028} \\
    \rowcolor{blue!20} PCP-MAE-Zero & - & - & \textbf{0.016} & \textbf{0.028} \\
    \bottomrule
  \end{tabular}
}
  \caption{\textbf{Masked Point Cloud Completion.} The table reports the $L_2$ Chamfer distance (\emph{lower is better}) between predicted masked points and ground truth on the test set of ShapeNet and procedurally synthesized 3D shapes. \textit{With Guidance}: center points of masked patches are added to mask tokens in the pretrained decoder, guiding masked point prediction during inference. \textit{Without Guidance}: \textit{Without Guidance}: no information from masked patches is available during inference.\vspace{-5px}}
  \label{tab:recon}
\end{table}

\subsection{Masked Point Cloud Completion}
\label{sec:recon}
Masked point cloud completion reconstructs missing regions of 3D point clouds as a self-supervised pretext task for learning 3D representations~\cite{pang2022masked} (see Fig.~\ref{fig:pipeline} and Sec.~\ref{sec:method}).

During pretraining, points are grouped into patches, with a subset of patches (60\% by default) randomly masked. Only visible patches are encoded, while masked patch centers can optionally guide the decoder (``with guidance'') or be omitted entirely (``without guidance''). After pretraining, models can reconstruct masked points even without such guidance. We quantitatively compare Point-MAE and Point-M2AE pretrained on ShapeNet (``-SN'') and procedural shapes (``-Zero'') in both guidance conditions, using the ShapeNet test split and 2,000 unseen synthetic shapes (see Tab.~\ref{tab:recon}). All methods perform slightly better on their in-domain data. Removing guidance significantly decreases performance across all methods, highlighting its importance during masked reconstruction. Notably, Point-MAE-Zero and Point-M2AE-Zero closely match or even surpass their SN counterparts in reconstructing synthetic shapes, and remain competitive on ShapeNet shapes despite the lack of semantic training signals. PCP-MAE is a special case since it predicts centers before decoding point cloud and we find PCP-MAE-SN and PCP-MAE-Zero achieve similar performances both on seen and unseen domains.

Fig.~\ref{fig:recon} further illustrates that procedural-only models (\eg, Point-MAE-Zero) effectively reconstruct familiar ShapeNet objects (\eg, airplane wings, chair legs) without semantic supervision, likely by exploiting geometric symmetries. Similarly, SN-pretrained models generalize effectively to synthetic shapes not encountered during pretraining. Overall, these findings from Fig.~\ref{fig:recon} and Tab.~\ref{tab:recon} underscore that masked autoencoding primarily captures geometric rather than semantic information, enabling robust reconstruction across domains.

\subsection{Scene-level 3D Understanding Tasks}
\label{sec:scene}

\begin{table}[t]
\centering
\resizebox{\columnwidth}{!}{
  \begin{tabular}{@{}l  c c c c c@{}}
    \toprule
    & \multicolumn{2}{c}{Semantic Seg.} & \multicolumn{3}{c}{Instance Segmentation} \\
    \cmidrule(lr){2-3} \cmidrule(lr){4-6}
    Methods & mIoU & mAcc & mAP & AP50 & AP25 \\
    \midrule
    MSC-scan~\cite{wu2023maskedscenecontrastscalable} & 73.85 & 81.80 & 39.75 & 60.51 & 76.49 \\
    \midrule
    MSC-Zero (1k) & 72.69 & 80.80 & 39.03 & 58.57 & 75.24 \\
    MSC-Zero (2k) & 73.86 & 82.03 & 40.81 & 62.28 & 76.26 \\
    \rowcolor{blue!20} MSC-Zero (4k) & \textbf{74.34} & \textbf{82.16} & \textbf{41.47} & \textbf{63.12} & \textbf{76.54} \\
    \bottomrule
  \end{tabular}
}
\vspace{-5px}
\caption{\textbf{Masked Scene Contrast Results.} Performance comparison between MSC-Scan and MSC-Zero with different amounts of pretraining data for semantic and instance segmentation tasks. \vspace{-5px}}
\label{tab:msc_results}
\end{table}

Given the effectiveness of procedural 3D programs for pretraining self-supervised learning (SSL) models on 3D objects, a natural question arises: \emph{Can procedural 3D programs similarly benefit SSL for 3D scenes?}  We adopt Masked Scene Contrast (MSC)~\cite{wu2023maskedscenecontrastscalable}, a popular SSL method for 3D scenes. We pretrain MSC on ScanNet~\cite{dai2017scannet} (1K scenes), commonly used for 3D scene SSL pretraining, and compare it against MSC pretrained on our procedurally generated scenes (denoted MSC-Zero). We conduct experiments with MSC-Zero using varying amounts of data (1K, 2K and 4K procedural scenes).  
MSC-Zero trained with 4K procedurally generated 3D scenes achieves outperforms MSC pretrained on ScanNet in both 3D semantic and instance segmentation tasks.
This demonstrates the effectiveness of procedurally generated data for 3D scene self-supervised learning and reinforces our findings from object-level 3D understanding tasks. Moreover, our results demonstrate that increasing the number of procedural scenes consistently improves performance across semantic and instance segmentation tasks.
We discuss the exact procedures of generating such data and implementation details in the supplementary materials.
\begin{figure*}[th!]
    \centering
    \begin{minipage}{0.4\textwidth}
        \centering
        \includegraphics[width=1\linewidth]{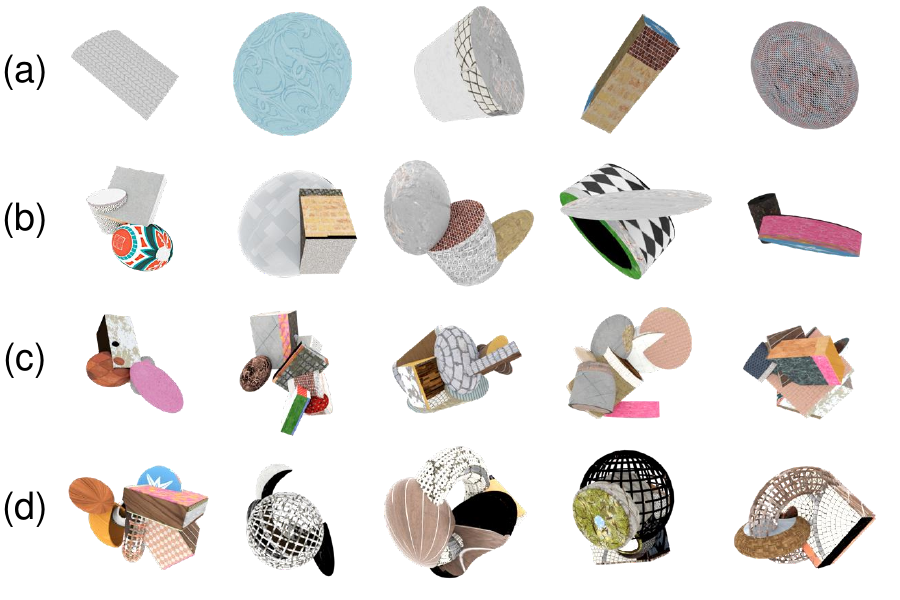}
    \end{minipage}%
    \hfill
    \begin{minipage}{0.55\textwidth}
        \centering
        \setlength{\tabcolsep}{1mm}
        \begin{tabular}{@{}l >{\centering\arraybackslash}p{20mm} >{\centering\arraybackslash}p{30mm}@{}}
            \toprule
            \textbf{Methods} & \textbf{Pre-train Loss} & \textbf{Downstream Accuracy} \\
            \midrule
            Scratch & -- & 77.24 \\ 
            Point-MAE-SN & 2.62 & 85.18 \\
            \midrule
            \textbf{Point-MAE-Zero} \\
            (a) Single Primitive & 3.17 & 83.93 \\
            \ \ (b) Multiple Primitives & 4.10 & 84.52 \\
            \ \ \ \ (c) Complex Primitives & 4.43 & 84.73 \\
            \ \ \ \ \ \  (d) Shape Augmentation & \textbf{5.28} & \textbf{85.46} \\
            \bottomrule
        \end{tabular}
    \end{minipage}
    \caption{\textbf{Impact of 3D Shape Complexity on Performance.} 
    \textbf{Left:} Examples of procedurally generated 3D shapes with increasing complexity, used for pretraining. \textit{Textures are shown for illustration purposes only;  in practice, only the surface points are used.} 
    \textbf{Right:} Comparison of pretraining masked point reconstruction loss (Eqn.~\ref{eq:recon})~\cite{pang2022masked} and downstream classification accuracy on the ScanObjectNN dataset~\cite{uy2019revisiting}. Each row in Point-MAE-Zero represents an incrementally compounded effect of increasing shape complexity and augmentation, with the highest accuracy achieved using shape augmentation.\vspace{-15px}}
    \label{fig:dataset_complexity}
\end{figure*}

\subsection{Analysis}
\label{sec:analysis}

\textbf{Complexity of Synthetic 3D shapes.} 
We examine how the geometric complexity of synthetic datasets impacts pretraining and downstream performance. We consider four progressively complex configurations: \textbf{(a) Single Primitive}: a single shape with affine transformations; \textbf{(b) Multiple Primitives ($\leq$3)}: up to three combined shapes; \textbf{(c) Complex Primitives ($\leq$9)}: up to nine combined shapes; \textbf{(d) Shape Augmentation}: further modified via boolean differences and wire-frame conversions.

Fig.~\ref{fig:dataset_complexity} displays samples from each configuration alongside quantitative comparisons of pretraining performance and downstream classification accuracy on PB-T50-RS, the most challenging variant of ScanObjectNN~\cite{uy2019revisiting}. As shape complexity increases, the pretraining task becomes more difficult, leading to higher reconstruction losses at the 300th training epoch. However, the downstream classification performance of Point-MAE-Zero improves. This underscores the importance of topological diversity in shapes for effective self-supervised point cloud representation learning.

We observe that the reconstruction loss on our dataset with single primitives (\ie, 3.17) is higher than on ShapeNet (\ie, 2.62) which consists of more diverse 3D shapes. We hypothesize that this is because ShapeNet is relatively smaller than our dataset (50K vs.~150K) and ShapeNet models are coordinate-aligned.

\begin{figure}[t!]
    \centering
    \includegraphics[width=0.95\linewidth]{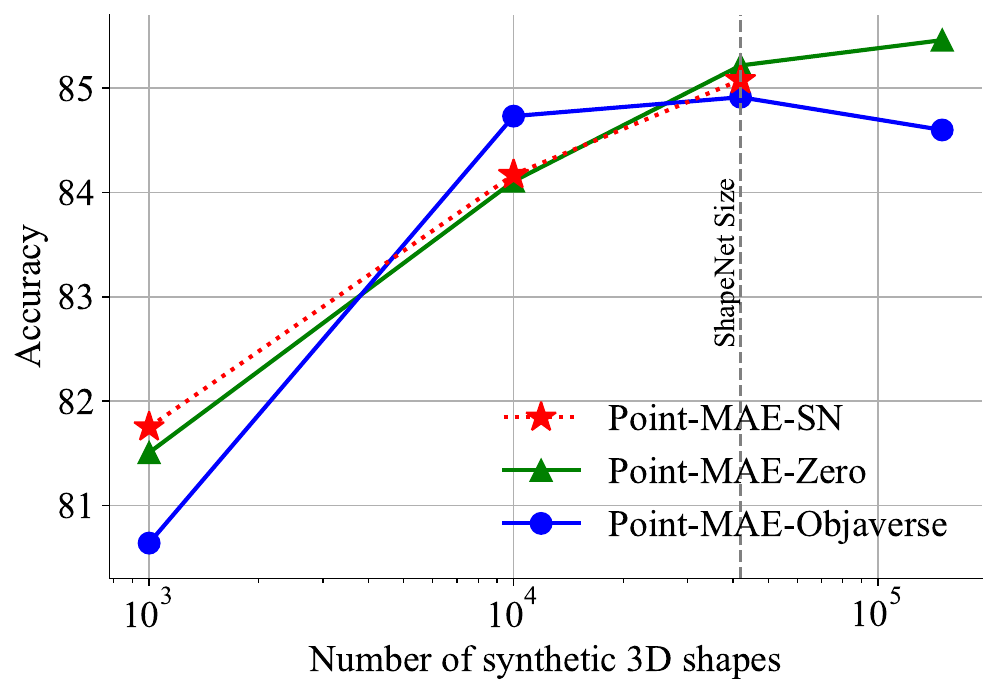}
    \vspace{-2mm}
    \caption{\textbf{Impact of pretraining dataset size.} We report the classification accuracy (\%) on the PB-T50-RS subset of ScanObjectNN~\cite{uy2019revisiting} as a function of the pretraining dataset size. \vspace{-10px}}
    \label{fig:datasetsize}
\end{figure}

\begin{figure}[ht!]
    \centering
    \includegraphics[width=\linewidth]{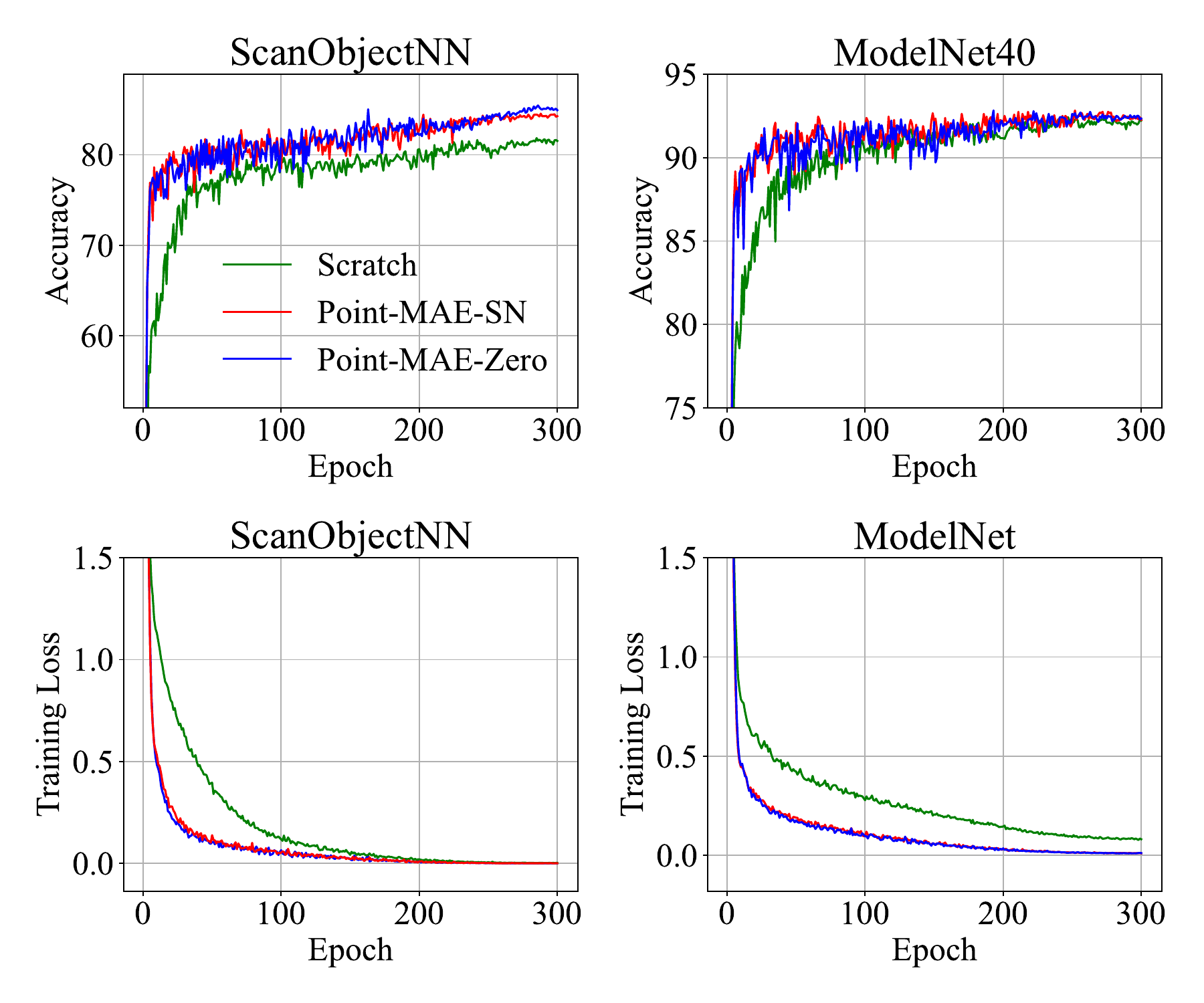}
    \vspace{-8mm}
\caption{\textbf{Learning curves in downstream tasks.} We present validation accuracy (top row) and training curves (bottom row) in object classification tasks on ScanObjectNN (left column) and ModelNet40 (right column). \vspace{-15px}}
    \label{fig:curve}
\end{figure}

\begin{figure*}[tp!]
    \centering
    \includegraphics[width=0.9\linewidth]{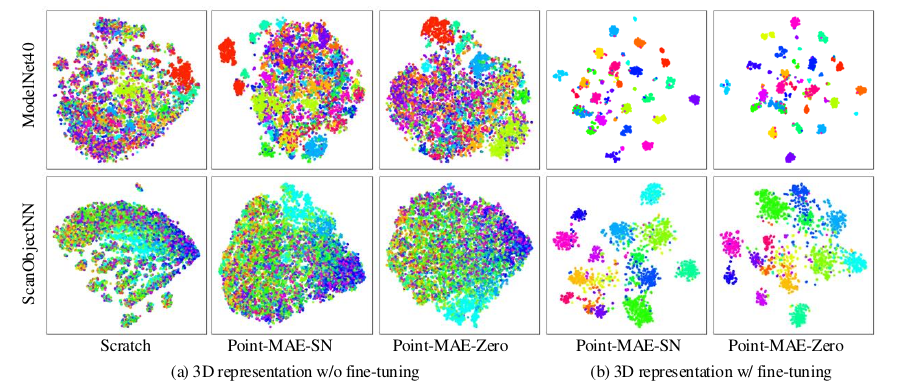}
    \vspace{-3mm}
    \caption{\textbf{t-SNE visualization of 3D shape representations.} \textbf{(a)} Shows representations from transformer encoders: Scratch, Point-MAE-SN (ShapeNet), and Point-MAE-Zero (procedural shapes).  \textbf{(b)} Displays fine-tuned representations for object classification on ModelNet40 (top) and ScanObjectNN (bottom). Each point represents a 3D shape while the color denotes the semantic categories. \vspace{-15px}}
    \label{fig:tsne}
\end{figure*}

\noindent
\textbf{Dataset Size and Comparison with Objaverse.} 
Fig.~\ref{fig:datasetsize} presents a scaling analysis of Point-MAE-Zero on the PB-T50-RS benchmark using procedurally generated data. When matched for dataset size, Point-MAE-Zero and Point-MAE-SN achieve comparable downstream performance, underscoring the viability of synthetic data as a substitute for curated datasets like ShapeNet. Importantly, we observe that performance further improves with increasing dataset size, despite the lack of semantics in the synthetic data. 

To contextualize our results, we compare against pretraining on randomly sampled subsets of Objaverse~\cite{deitke2023objaverse}, matched in scale. 
While models pretrained on Objaverse exhibit strong performance at smaller scales (\eg, $10^3$ or $10^4$ shapes), their downstream performance plateaus—and eventually declines—as more shapes are added. We pretrain and finetune Point-MAE-Objaverse 5 times on PB-T50-RS and we report the best performance.
We hypothesize that this trend is due to many randomly sampled shapes from Objaverse having simple geometry (\eg, avatars, boxes, cups), which makes the pretraining task too easy and limits the ability to learn robust 3D representations.
In contrast, our procedurally generated data, enriched through augmentation and shape composition, consistently improves performance as the dataset scales, demonstrating its effectiveness and scalability for point cloud self-supervised learning.

\noindent
\textbf{Efficiency of Transfer Learning.} Fig.~\ref{fig:curve} shows the learning curves for training from scratch, Point-MAE-SN, and Point-MAE-Zero on shape classification tasks in the transfer learning setting. Both Point-MAE-SN and Point-MAE-Zero demonstrate faster training convergence and higher accuracy compared to training from scratch, consistently across both ModelNet40 and ScanObjectNN benchmarks.

\noindent
\textbf{t-SNE Visualization. }Fig.~\ref{fig:tsne} visualizes the distribution of 3D shape representations from Point-MAE-SN and Point-MAE-Zero via t-SNE~\cite{van2008visualizing}, before and after fine-tuning on specific downstream tasks. It also includes representations from a randomly initialized neural network as a reference.

First, compared to the representations from scratch, both Point-MAE-SN and Point-MAE-Zero demonstrate \textbf{\emph{visually improved separation}} between different categories in the latent space. For example, this is evident in the red and light blue clusters on ModelNet40  and the blue and light blue clusters on ScanObjectNN. This highlights the effectiveness of masked auto-encoding for self-supervised 3D learning.

Second, when comparing representations after fine-tuning, both Point-MAE-SN and Point-MAE-Zero show \textbf{\emph{much less clear separation}} between categories in the latent space. This raises the question of whether high-level semantic features are truly learned through the masked autoencoding pretraining scheme.

Finally, the t-SNE visualization reveals structural similarities between Point-MAE-Zero and Point-MAE-SN. Most categories \textbf{\emph{lack clear separation}} in both models, except for the red and light blue clusters on ModelNet40 and the blue and light blue clusters on ScanObjectNN. This suggests that Point-MAE-Zero and Point-MAE-SN may have learned similar 3D representations, despite differences in the domains of their pretraining datasets. 
We provide more in-depth analysis in the supplementary material.
\section{Discussion}
\label{sec:Discussion}

In this work, we propose to learn 3D representations from synthetic data automatically generated using procedural 3D programs. 
We conduct an comprehensive empirical analysis of existing 3D SSLs and perform extensive comparisons with learning from well-curated, semantically meaningful 3D datasets.

We demonstrate that learning with procedural 3D programs performs comparably to learning from recognizable 3D models, despite the lack of semantic content in synthetic data. 
Our experiments highlights the importance of geometric complexity and dataset size in synthetic datasets for effective 3D representation learning.
Our analysis further reveals that existing 3D SSLs primarily learns geometric structures (\eg, symmetry) rather than high-level semantics. 

This work has several limitations. For example, due to limited computational resources, we were unable to further scale up our experiments, such as by increasing the dataset size or conducting more detailed ablation studies on procedural 3D generation. 
Additionally, our findings may be influenced by potential biases in visualization tools (\eg, t-SNE) or benchmarks (\eg, data distribution and evaluation protocols).
Furthermore, in 3D vision, the distinction between geometric structures and semantics remains an open question, as well-stated by Xie~\etal~\cite{xie2024lrm}.
This work also does not provide any novel representation learning method.
Nevertheless, we hope our findings will inspire further exploration into self-supervised 3D representation learning.

\section{Acknowledgment}

The authors acknowledge the Adobe Research Gift, the University of Virginia Research Computing and Data Analytics Center, Advanced Micro Devices AI and HPC Cluster Program, Advanced Cyberinfrastructure Coordination Ecosystem: Services \& Support (ACCESS) program, and National Artificial Intelligence Research Resource (NAIRR) Pilot for computational resources, including the Anvil supercomputer (National Science Foundation award OAC 2005632) at Purdue University and the Delta and DeltaAI advanced computing resources (National Science Foundation award OAC 2005572).
{
    \small
    \bibliographystyle{ieeenat_fullname}
    \bibliography{main}
}
\clearpage
\appendix
\setcounter{page}{1}
\maketitlesupplementary

We provide more implementation details, additional experimental results, and visualizations in this supplementary material. 
In Sec.~\ref{sec:scratch-baseline}, we present our training-from-scratch baseline, which surpasses the results reported in Point-MAE \cite{pang2022masked}. In Sec.~\ref{sec:rigorous}, we introduce a more rigorous evaluation protocol using a dedicated validation set instead of relying on the test set for validation. Moreover, in Sec.~\ref{sec:linear}, we provide linear probing results on ModelNet40 and three ScanObjectNN variants to compare the performance of Point-MAE-SN and Point-MAE-Zero. 
Additional visualizations are provided in Sec.\ref{sec:additional_vis}. Details on generating scene-level procedural 3D programs are in Sec.\ref{sec:generate_detail}. Lastly, implementation details for 3D object SSLs and 3D scene SSLs are presented in Sec.~\ref{sec:detail}.

\section{Training-from-Scratch Baseline}
\label{sec:scratch-baseline}

For the training-from-scratch baseline, we report the results in the supplementary material for completeness. In the main text, we referenced the scores reported in the original Point-MAE paper for this baseline. However, our experiments suggest that training from scratch in our setup produced higher scores than those reported in Point-MAE~\cite{pang2022masked} and Point-BERT~\cite{yu2022point}. The results from our training-from-scratch baseline are presented in this section, as shown in Tab.~\ref{tab:better-train-scratch_object_classification}, Tab.~\ref{tab:fewshot-better-scratch}, and Tab.~\ref{tab:partseg-better-scratch}.

We follow the same evaluation protocol as Point-MAE~\cite{pang2022masked}. While the results we obtained from the training-from-scratch baseline consistently surpass the previously reported scores, there remains a significant gap between the performance of the training-from-scratch baseline and the pre-trained methods. This underscores the effectiveness of pre-training in enhancing model performance. Furthermore, pre-trained methods demonstrate much faster convergence compared to the training-from-scratch baseline.

\begin{table}[ht!]
\resizebox{\columnwidth}{!}{
  \centering
  \begin{tabular}{@{} l c c c c@{}}
    \toprule
    Methods  & ModelNet40 & OBJ-BG & OBJ-ONLY & PB-T50-RS\\
    \midrule
    Scratch~\cite{yu2022point} & 91.4 & 79.86 & 80.55 & 77.24 \\
    Scratch\textsuperscript{*} & 93.4 & 87.44 & 82.03 & 81.99 \\
    Point-MAE-SN~\cite{pang2022masked} & \textbf{93.8} & 90.02 & 88.29 & 85.18 \\  
    Point-MAE-Zero & 93.0 & \textbf{90.36} & \textbf{88.64} & \textbf{85.46} \\
    \bottomrule
  \end{tabular}
}
\caption{\textbf{Object Classification.} Classification accuracy (\%) on ModelNet40 and three ScanObjectNN variants under the revised evaluation setup (\emph{Higher is better}). Note Scratch\textsuperscript{*} indicates baseline method's results we obtained.}
\label{tab:better-train-scratch_object_classification}
\end{table}

\begin{table}[ht!]
\resizebox{\columnwidth}{!}{
  \centering
  \begin{tabular}{@{}l c c c c@{}}
    \toprule
     Methods & 5w/10s & 5w/20s & 10w/10s & 10w/20s \\
    \midrule
    Scratch~\cite{yu2022point} & 87.8$\pm$5.2 & 93.3$\pm$4.3  & 84.6$\pm$5.5  & 89.4$\pm$6.3  \\
    Scratch\textsuperscript{*}  & 92.7$\pm$3.5 & 95.5$\pm$3.1 & 88.5$\pm$5.1 & 92.0$\pm$4.5\\
    Point-MAE-SN~\cite{pang2022masked} &  96.3$\pm$2.5 &  \textbf{97.8$\pm$1.8} &  \textbf{92.6$\pm$4.1} & 95.0$\pm$3.0\\
    Point-MAE-Zero & \textbf{96.6$\pm$2.2} & 97.6$\pm$1.4 & 91.9$\pm$4.2 & \textbf{95.2$\pm$3.0} \\
    \bottomrule
  \end{tabular}
  }
  \caption{\textbf{Few-shot classification on ModelNet40.} We evaluate performance on four \(n\)-way, \(m\)-shot configurations. For example, 5w/10s denotes a 5-way, 10-shot classification task. The table reports the mean classification accuracy (\%) and standard deviation across 10 independent runs for each configuration. \textbf{Top}: Results from existing methods for comparison. \textbf{Bottom}: Comparison with our baseline methods. Note Scratch\textsuperscript{*} indicates baseline method's results we obtained.}
  \label{tab:fewshot-better-scratch}
\end{table}

\begin{table}[ht!]
\centering
\resizebox{\columnwidth}{!}{
  \begin{tabular}{@{}l | c | c c c c c c c c@{}}
    \toprule   
    Methods & mIoU\textsubscript{I} & aero & bag & cap & car & chair & earphone & guitar & knife \\
    \midrule
    Scratch~\cite{yu2022point} & 85.1 & 82.9 & \textbf{85.4} & 87.7 & 78.8 & 90.5 & \textbf{80.8} & 91.1 & \textbf{87.7} \\
    Scratch\textsuperscript{*} & 84.0 & 84.3 & 83.1 & \textbf{89.1} & 80.6 & 91.2 & 74.5 & \textbf{92.1} & 87.3 \\
    Point-MAE-SN~\cite{pang2022masked} & \textbf{86.1} & 84.3 & 85.0 & 88.3 & 80.5 & 91.3 & 78.5 & \textbf{92.1} & 87.4 \\  
    Point-MAE-Zero & \textbf{86.1} & \textbf{85.0} & 84.2 & 88.9 & \textbf{81.5} & \textbf{91.6} & 76.9 & \textbf{92.1} & 87.6 \\
    \bottomrule
  \end{tabular}
}

\vspace{2mm}

\resizebox{\columnwidth}{!}{
  \begin{tabular}{@{}l | c c c c c c c c c@{}}
    \toprule
    Methods & lamp & laptop & motor & mug & pistol & rocket & skateboard & table \\
    \midrule
    Scratch~\cite{yu2022point} & 85.3 & 95.6 & 73.9 & \textbf{94.9} & 83.5 & 61.2 & 74.9 & 80.6 \\
    Scratch\textsuperscript{*} & 85.1 & 95.9 & 74.3 & 94.8 & 84.3 & 61.1 & 76.2 & 80.9  \\
    Point-MAE-SN~\cite{pang2022masked} & \textbf{86.1} & \textbf{96.1} & 75.2 & 94.6 & 84.7 & 63.5 & 77.1 & \textbf{82.4} \\  
    Point-MAE-Zero & 86.0 & 96.0 & \textbf{77.8} & 94.8 & \textbf{85.3} & \textbf{64.7} & \textbf{77.3} & 81.4 \\
    \bottomrule
  \end{tabular}
}
\caption{\textbf{Part Segmentation Results.} We report the mean Intersection over Union (IoU) across all instances (mIoU\textsubscript{I}) and the IoU (\%) for each category on the ShapeNetPart benchmark (\emph{higher values indicate better performance}). Note Scratch$\textsuperscript{*}$ indicates baseline method's results we obtained.}
\label{tab:partseg-better-scratch}
\end{table}

\section{More Rigorous Evaluation}
\label{sec:rigorous}
In prior works~\cite{yu2022point, pang2022masked}, the validation set was identical to the test set. The model was evaluated on the test set after every epoch, with the best-performing result selected. Such practices can artificially inflate performance metrics and fail to accurately reflect the model’s ability to generalize to unseen data. 
While we followed this setup in the main text for fair comparisons, we also performed more rigorous evaluations using a dedicated validation set, which has no overlap with either the training set or the test sets.
Specifically, we set aside 20\% of the original test set as our validation set, leaving the remaining 80\% as the new test set. 
We then report the test performance with checkpoints selected based on the validation accuracy. 

As presented in Tab.~\ref{tab:rigorous_eval}, our new evaluation results are consistent with these reported in the main text --- both Point-MAE-SN and Point-MAE-Zero outperform the training-from-scratch baseline across all four object classification tasks, while Point-MAE-Zero performs on par with Point-MAE-SN.  
The performance gap between models with pretraining and models trained from scratch is particularly pronounced in the most challenging experiment, PB-T50-RS, compared to the other three classification tasks. 

On ModelNet40, we observe that training-from-scratch and pre-trained methods achieve similar performance, though the results may vary across different runs. For example, in another run with a different random seed, training-from-scratch achieve 92.1\% accuracy, which is lower than both Point-MAE-SN and Point-MAE-Zero.
However, in all our experiments, pre-trained methods consistently converge significantly faster than training-from-scratch as shown in Fig.~\ref{fig:curve} in the main text. 

\begin{table}[H]
\resizebox{\columnwidth}{!}{
  \centering
  \begin{tabular}{@{} l c c c c@{}}
    \toprule
    Methods  & ModelNet40 & OBJ-BG & OBJ-ONLY & PB-T50-RS\\
    \midrule
    Scratch\textsuperscript{*}~\cite{yu2022point} & \textbf{92.95} & 84.19 & 86.94 & 80.92 \\
    Point-MAE-SN~\cite{pang2022masked} & 92.22 & \textbf{88.32} & 87.97 & 83.83 \\ 
    Point-MAE-Zero & 92.87 & \textbf{88.32} & \textbf{88.31} & \textbf{84.73} \\
    \bottomrule
  \end{tabular}
}
\caption{\textbf{Object Classification.} Classification accuracy (\%) on ModelNet40 and three ScanObjectNN variants under the revised evaluation setup. The original test set was split to create a new test set and validation set, and all models were re-evaluated using the updated splits. (\emph{Higher is better}). Note Scratch\textsuperscript{*} indicates baseline method's results we obtained and we do not apply voting in these experiments. \vspace{-10pt}}
\label{tab:rigorous_eval}
\end{table}

\section{Linear Probing}
\label{sec:linear}

In line with Point-BERT~\cite{yu2022point} and Point-MAE~\cite{pang2022masked}, we primarily report the performance of pre-trained methods in a transfer learning and few-shot learning setting in the main text, where the pre-trained model is fine-tuned with task-specific supervision. 
However, a common practice for benchmarking self-supervised learning methods is the linear probing, which trains a single linear layer while keeping the  pretrained network backbone frozen. 
Below we provide details of our experimental setup and results.

\paragraph{Experimental Setup.}

Instead of fully fine-tuning the pre-trained model, we freeze the model’s weights and train a single linear layer for the target task. 

We include a training-from-scratch baseline, where we freeze the randomly initialized weights and train only a single linear layer, providing a point of comparison for the pre-trained models.

\begin{table}[H]
\resizebox{\columnwidth}{!}{
  \centering
  \begin{tabular}{@{} l c c c c@{}}
    \toprule
    Methods  & ModelNet40 & OBJ-BG & OBJ-ONLY & PB-T50-RS\\
    \midrule
    Scratch\textsuperscript{*} & 84.16 & 62.65 & 66.09 & 56.80 \\
    Point-MAE-SN~\cite{pang2022masked} & \textbf{90.56} & \textbf{78.83} & \textbf{81.41} & \textbf{68.22} \\  
    Point-MAE-Zero & 89.30 & 76.24 & 78.83 & 67.87 \\
    \bottomrule
  \end{tabular}
}
\caption{\textbf{Object Classification.} Classification accuracy (\%) on ModelNet40 and three ScanObjectNN variants under the revised evaluation setup (\emph{Higher is better}). Note Scratch\textsuperscript{*} indicates baseline method's results we obtained. \vspace{-10pt}}
\label{tab:linear_probing}
\end{table}

\paragraph{Results.}

We present linear probing results for object classification in Tab.~\ref{tab:linear_probing}. The gap between pre-trained models and the training-from-scratch baseline is noticeably larger in this setting. Interestingly, Point-MAE-SN outperforms Point-MAE-Zero in object classification under the linear probing setup, suggesting that semantically meaningful data may enable models to achieve a better understanding of 3D structures. However, the performance gap between Point-MAE-SN and Point-MAE-Zero remains relatively small. 
Based on the results in Tab.~\ref{tab:fine-tuning} in the main text, Point-MAE-SN appears to be a more effective choice for transfer learning tasks. We encourage future research to explore improved learning algorithms to take full advantage of data generated from procedure 3D programs.

\section{Additional Visualization}
\label{sec:additional_vis}

In this section, we present additional qualitative comparisons between Point-MAE-SN and Point-MAE-Zero for Masked Point Cloud Completion under both \textit{guided} and \textit{unguided} settings. Additionally, we provide t-SNE visualizations for Point-MAE-Zero and Point-MAE-SN to further investigate its representational capabilities, focusing on whether it can effectively distinguish between different primitives.

\subsection{Masked Point Cloud Completion}
\label{sec:recon_additional}

\begin{figure*}[t!]
    \centering
        \includegraphics[width=0.88\linewidth]{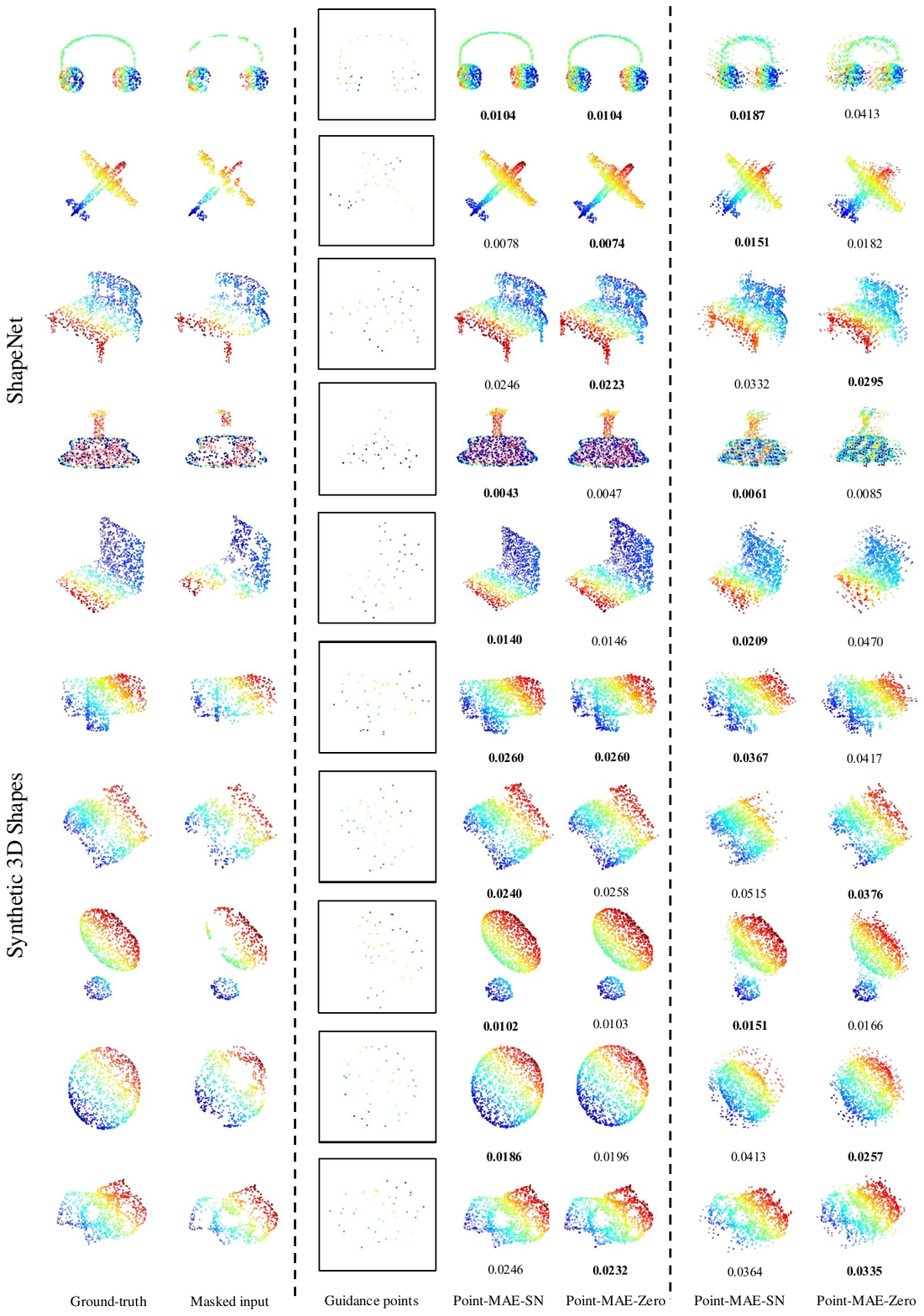}
    \vspace{-6mm}
    \caption{\textbf{Masked Point Cloud Completion.} This figure visualizes shape completion results with Point-MAE-SN and Point-MAE-Zero on the test split of ShapeNet and procedurally synthesized 3D shapes. 
    \textbf{Left}: Ground truth 3D point clouds and masked inputs with a 60\% mask ratio. 
    \textbf{Middle}: Shape completion results using the centers of masked input patches as guidance, following the training setup of Point-MAE~\cite{pang2022masked}.
    \textbf{Right}: Point cloud reconstructions without any guidance points. 
    The $L_2$ Chamfer distance (\emph{lower is better}) between the predicted 3D point clouds and the ground truth is displayed below each reconstruction.
}
    \label{fig:recon_full_page}
\end{figure*}

Additional qualitative results for masked point cloud completion, both guided and unguided, are shown in Figure~\ref{fig:recon_full_page}.

\subsection{More t-SNE Visualizations}
\begin{figure}[H]
    \centering
    \includegraphics[width=\linewidth]{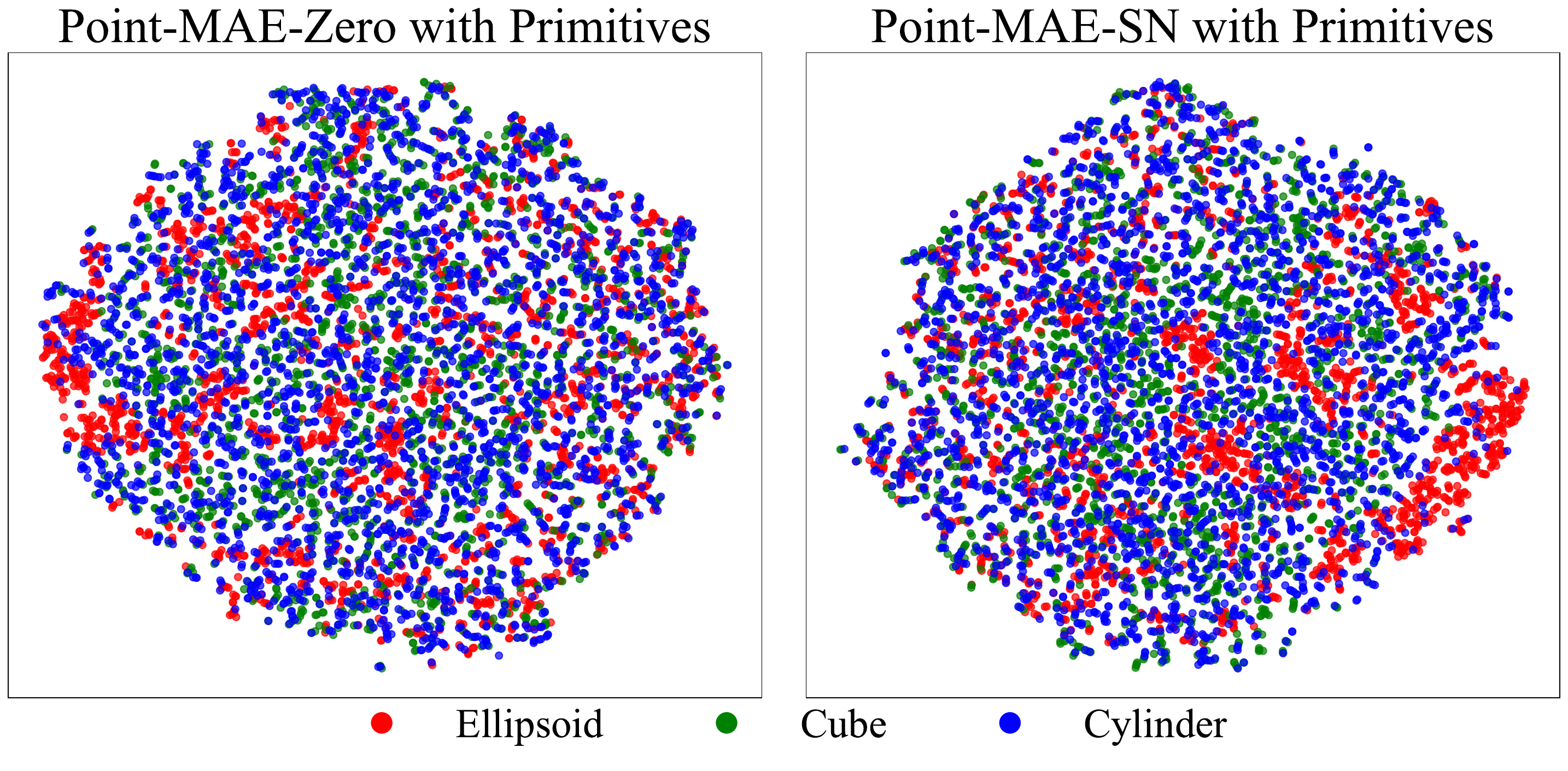}
    \caption{\textbf{t-SNE Visualization.} We visualize features extracted by Point-MAE-Zero (left) and Point-MAE-SN (right) for three primitive shapes --- Ellipsoid, Cube, and Cylinder. \vspace{-10pt}}
    \label{fig:t-sne-primitives}
\end{figure}
\paragraph{Point-MAE on Primitives}
Fig.~\ref{fig:t-sne-primitives} presents t-SNE visualizations of features extracted by Point-MAE-Zero and Point-MAE-SN for three primitives: ellipsoid, cube, and cylinder.
Interestingly, the notable structural differences among these primitives are not reflected in the latent space of either Point-MAE-SN or Point-MAE-Zero. 
We hypothesize that this occurs because the learned 3D representations from both models primarily capture local structures rather than global shapes.

\section{Details on Generating Scene-Level Procedural 3D Programs}
\label{sec:generate_detail}

We follow MegaSynth's ~\cite{jiang2024megasynth} generation pipeline.  First, the pipeline generates a scene floor plan. Each 3D box represents a different shape, with distinct colors indicating various object types. This defines the spatial structure of the scene. Next, we creates 3D objects by combining primitive shapes such as cubes and spheres. These objects undergo geometry augmentations, including scaling, deformation, and boolean operations, to introduce variations. Since we do not need texture and lighting, we omit the rest of the procedurals. Instead we sample points from each shape and randomly sample points from the layout in order to represent walls, ceilings and floors. We will release code to further assist the reproducibility of our results.

\section{Implementation details.}
\label{sec:detail}
We closely follow each baseline's open‐source configuration. Similar to prior work for 3D shape pretraining, we sample each point cloud to \(p = 1024\) points and divide it into \(n = 64\) patches, with each patch containing \(k = 32\) points via the KNN algorithm. The autoencoder consists of an encoder with 12 Transformer blocks and a decoder with 4 Transformer blocks, each block having a 384‐dimensional hidden size and 6 attention heads. During pretraining, we randomly sample 1024 points per shape and apply standard random scaling and translation. We train for 300 epochs using the AdamW optimizer~\cite{loshchilov2016sgdr} with a cosine decay schedule~\cite{loshchilov2017decoupled}, an initial learning rate of 0.001, weight decay of 0.05, and a batch size of 128.

For scene-level SSL MSC, we use SparseUNet34 as the backbone with a hierarchical encoder-decoder structure. The encoder consists of depths \([2, 3, 4, 6]\) and channels \([32, 64, 128, 256]\), while the decoder follows depths \([2, 2, 2]\) with channels \([256, 128, 64, 64]\). A kernel size of 3 is used in both encoding and decoding, with a pooling stride of \([2, 2, 2, 2]\). The pre-training phase employs SGD~\cite{ruder2016overview} with a cosine decay scheduler~\cite{loshchilov2017decoupled}, an initial learning rate of 0.1, a weight decay of \(1e^{-4}\), and a momentum of 0.8. We use a batch size of $32$ and train on ScanNet for 600 epochs, with 6 warmup epochs. For fine-tuning, we use a similar SGD~\cite{ruder2016overview} with cosine decay setup~\cite{loshchilov2017decoupled}, but with a learning rate of 0.05, weight decay of \(1e^{-4}\), and momentum of 0.9. The batch size is 48, with 40 warmup epochs, and the model is trained for 600 epochs.
We will release code to further assist the reproducibility of our results.

\end{document}